%% file: GradientBackpropGATv2.tex
\documentclass[journal, onecolumn]{IEEEtran}  

\usepackage{amsmath} 
\usepackage{amssymb}  
\usepackage{bbm}

\usepackage{cite}
\usepackage{algorithmic}
\usepackage{graphicx}
\usepackage{textcomp}
\usepackage[svgnames]{xcolor}
\usepackage{bm}
\usepackage{multirow}
\usepackage{siunitx}
\usepackage{tikz, pgfplots}
\usepackage{mathtools}
\usepackage{verbatim}
\usepackage{babel}
\usepackage{filecontents}
\usepackage{pgfplotstable}
\usepgfplotslibrary{colorbrewer}
\pgfplotsset{compat = 1.15, cycle list/Set1-8} 
\usetikzlibrary{pgfplots.statistics, pgfplots.colorbrewer} 
\usepackage{filecontents}
\usepackage{float} 
\usepackage{graphicx}
\usepackage{caption}
\usepackage{subcaption}

\usepackage{ifthen}
\usepackage{pgfplots}
\usepackage{subcaption}
\graphicspath{{./fig/}}
\usetikzlibrary{arrows, calc,intersections, shapes}
\usetikzlibrary{arrows.meta}
\usetikzlibrary{decorations.pathmorphing}
\usetikzlibrary{decorations.text}
\usetikzlibrary{decorations.pathreplacing}
\usetikzlibrary{hobby,decorations.markings}
\usetikzlibrary{positioning}
\usetikzlibrary{calc}
\tikzset{%
	every neuron/.style={
		circle,
		draw,
		minimum size=0.15cm
	},
	neuron missing/.style={
		draw=none, 
		scale=1,
		text height=1,
		execute at begin node=\color{black}$\vdots$
	},
}

\usepackage[switch]{lineno}  
\usepackage[hyphens]{url}  
\def\BibTeX{{\rm B\kern-.05em{\sc i\kern-.025em b}\kern-.08em
		T\kern-.1667em\lower.7ex\hbox{E}\kern-.125emX}}

\makeatletter
\newcommand{\newlineauthors}{%
\end{@IEEEauthorhalign}\hfill\mbox{}\par
\mbox{}\hfill\begin{@IEEEauthorhalign}
}
\makeatother
\begin{document}
\title{Gradient Derivation for Learnable Parameters in Graph Attention Networks}
\author{
	Marion Neumeier\textsuperscript{\rm 1}*,
	Andreas Tollkühn\textsuperscript{\rm 2},
	Sebastian Dorn\textsuperscript{\rm 2},
	Michael Botsch\textsuperscript{\rm 1} and 
	Wolfgang Utschick\textsuperscript{\rm 3}\\	
	\thanks{*We appreciate the funding of this work by AUDI AG.}
	\thanks{\textsuperscript{\rm 1} CARISSMA Institute of Automated Driving, Technische Hochschule Ingolstadt, 85049 Ingolstadt, Germany {\tt\small firstname.lastname@thi.de}}%
	\thanks{\textsuperscript{\rm 2} AUDI AG, 85057 Ingolstadt, Germany {\tt\small firstname.lastname@audi.de}}
	\thanks{\textsuperscript{\rm 3} School of Computation, Information and Technology, Technical University of Munich, 80333 Munich, Germany {\tt\small utschick@tum.de}}
}
\maketitle
\thispagestyle{empty}
\begin{abstract}
This work provides a comprehensive derivation of the parameter gradients for GATv2\cite{GATv2}, a widely used implementation of Graph Attention Networks (GATs). GATs have proven to be powerful frameworks for processing graph-structured data and, hence, have been used in a range of applications. However, the achieved performance by these attempts has been found to be inconsistent across different datasets and the reasons for this remains an open research question. As the gradient flow provides valuable insights into the training dynamics of statistically learning models, this work obtains the gradients for the trainable model parameters of GATv2. The gradient derivations supplement the efforts of \cite{Neumeier.2023}, where potential pitfalls of GATv2 are investigated.
%
\end{abstract}
\section{Introduction}
Over the past years, Graph Attention Networks (GATs) \cite{Velickovic.30.10.2017}\cite{GATv2} have been gaining increasing popularity for representation learning on graph-structured data. GATs update node features by aggregating the representations of neighboring nodes with a weighted sum based on attention scores assigned to each neighbor. The attention mechanism can improve a network’s robustness and performance as it enables attending to relevant nodes only in graph-structured data\cite{understandGAT}. Similar to conventional Neural Networks, GATs are commonly trained using error backpropagation and  gradient descent. By computing the gradient of the loss function with respect to the model's weights, backpropagation provides a way to determine how each weight affects the output of the network and how to adjust those weights to minimize the loss. It represents a systematic way to propagate the error backward and determine the gradient for each network parameter. This gradient is then used to update the weights using an optimization algorithm such as gradient descent. Through repeatedly applying this process, the network is optimized and learns to decode relevant information from graph-structured data. Consequently, a robust learning characteristic is highly dependent on a backward pass that allows a consistent gradient flow. The choice of activation function for neural networks can greatly impact the stability of learning behavior. For instance, one common problem with certain activation functions, such as ReLU, is the occurrence of "dying neurons". This refers to neurons that become unresponsive due to the activation function saturating, i.\,e. its derivative during backpropagation is zero or nearly zero. In such cases, the network may fail to converge or suffer from slow training.

To gain understanding of the training behavior of GATv2\cite{GATv2}, this work derives the gradients for its network parameters. This study is an addition to \cite{Neumeier.2023}, in which potential drawbacks and issues of GATv2 are analyzed. Certain hypotheses of \cite{Neumeier.2023} are evidenced and supported upon the outcome of this study.
\section{Preliminaries}
In this work, vectors are denoted as bold lowercase letters and matrices as bold capital letters.
\subsection{Graph definition} Let $\mathcal{G} = (\mathcal{V}, \mathcal{E})$ be a graph composed of nodes $\mathcal{V}=\{1, \dots, n\}$ and edges $\mathcal{E} \subseteq \mathcal{V} \times \mathcal{V}$. An edge from a node $j$ to a node $i$ is represented by $(i,j) \in \mathcal{E}$. If all edges are bidirectional the graph is denoted as undirected; and directed if otherwise. The graph $\mathcal{G}$ can be represented through its adjacency matrix $\bm{A} = \{0, 1\}^{|\mathcal{V}|\times|\mathcal{V}|}$. If the graph is weighted, an additional weight matrix $\bm{W}~\in~\mathbb{R}^{|\mathcal{V}|\times|\mathcal{V}|}$, indicating the weighting of each edge, can be defined.
\subsection{Graph Attention Networks} 
\begin{figure}[h]
	\centering
	\vspace{-3pt}
	\includegraphics{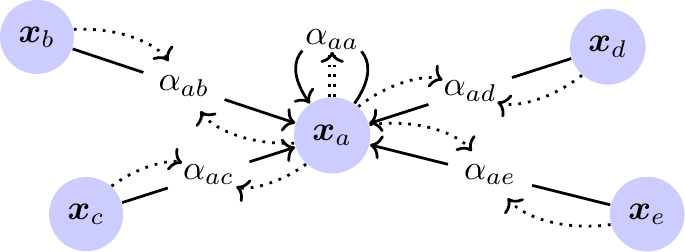}
	\caption{Concept of attentional message passing on graphs.\cite{Neumeier.2023}}
	\label{fig:GAT}
\end{figure}

GATs are realizations of Graph Neural Networks (GNNs) operating on the concept of message-passing. In GATs, the features of the neighboring nodes are aggregated by computing attention scores $\alpha_{ij}$ for every edge~$\left(i,j\right)$. Initially, each node  $i \in \mathcal{V}$ of the graph structure is parameterized based on the corresponding data features \mbox{$\bm{h}_i \in \mathbb{R}^H$.} During the message-passing process, node $i$ computes a weighted sum of the features of its neighbors $j \in \mathcal{N}_i$, where the weights are the attention coefficients. Subsequently, the aggregated information of the neighboring nodes is combined with the current features of node $i$ to update its node representation. In Fig.~\ref{fig:GAT}, the conceptual idea of GATs to perform a weighted sum over the neighboring nodes $j \in \mathcal{N}_i$ based on the attention scores $\alpha_{ij}$ is shown. 

There are different ways to compute attention scores. A very popular approach is GATv2\cite{GATv2}, where the network updates the features of node $i$ as shown in \mbox{Eq. \ref{eq:Nscore}-\ref{eq:GATV2UPDATE}}. The equations correspond to the default implementation of GATv2 in the PyTorch Geometric framework~\cite{pytorchGATv2CONV}. 
\begin{align}
e(\tilde{\bm{h}}_i, \tilde{\bm{h}}_j) &= \bm{a}^\mathrm{T}\text{LeakyReLU}\left({\boldsymbol{\Theta}}_R\tilde{\bm{h}}_i +{\boldsymbol{\Theta}}_L \tilde{\bm{h}}_j \right) \label{eq:Nscore}\\
\alpha_{ij} &= \text{softmax}_j(e (\bm{\tilde{h}}_i, \bm{\tilde{h}}_j) ) \label{eq:sftmax}\\
\bm{h}_i' &= \bm{b}  + \sum_{j \in \mathcal{N}_i} \alpha_{ij} {\boldsymbol{\Theta}}_L\tilde{\bm{h}}_j
\label{eq:GATV2UPDATE}
\end{align}
The attention scores are determined by the scoring function (Eq.~\ref{eq:Nscore}), where $\bm{a}\in \mathbb{R}^{D}$,  $\boldsymbol{\Theta}_p \in \mathbb{R}^{D \times (H+1)}$ for \mbox{$p\in\{R,L\}$} are learned and $\tilde{\bm{h}}_q = [1, \bm{h}_q^\mathrm{T}]^\mathrm{T}$ for $q\in\{i,j\}$ are node representations.
To obtain the attention scores, the embeddings of the central node ${\bm{h}}_i$ and its neighbors ${\bm{h}}_j$ are transformed with the weight matrices $\boldsymbol{\Theta}_R, \boldsymbol{\Theta}_L$ and then passed through the LeakyReLU activation. By computing the dot product of the resulting representation and $\bm{a}$, a scalar attention score for the neighboring node $j$ is determined. The resulting scores of the scoring function $e(\tilde{\bm{h}}_i, \tilde{\bm{h}}_j)$ for all neighboring nodes $j$ are normalized in Eq.~\ref{eq:sftmax} using softmax such that $\sum_j \alpha_{ij} = 1$. Subsequently, the normalized attention scores are used to update the feature representation by computing a weighted sum as described in Eq. \ref{eq:GATV2UPDATE}, where $\bm{b} \in \mathbb{R}^{D}$ is a learnable parameter. 
\subsection{Gradient computation using the Jacobian matrix}
The Jacobian matrix is a matrix of all first-order partial derivatives of a vector-valued function. In the context of machine learning, the Jacobian matrix can be used to determine the gradients of the loss function with respect to the network parameters. 
During forward propagation in a neural network, the output of each layer is calculated based on the input from the previous layer and layer's parameters. Thereby, the input data is propagated through the network layer by layer to the output layer. During backpropagation, the gradients of the loss function with respect to the output of each layer are propagated backwards through the network to update the weights and biases.

To calculate the gradients of the network parameters, the Jacobian matrix of the output of each layer with respect to the input to that layer is determined. If a layer $f : \mathbb{R}_n\rightarrow \mathbb{R}_m$ is a differentiable function computing $\bm{y}=f(\bm{x})$, then the Jacobian matrix $\mathbf{J}$ of $f$ is a $m \times n$ matrix such that
\begin{equation}
\mathbf{J} =
\frac{\partial (\text{vec}\{f(\bm{x})\})}{\partial (\text{vec}\{\bm{x}\})^\textrm{T}}=
\begin{bmatrix}
\frac{\partial f(\bm{x})}{\partial x_1}\quad
\cdots \quad
\frac{\partial f(\bm{x})}{\partial x_n}
\end{bmatrix} =
\begin{bmatrix}
\frac{\partial f_1}{\partial x_1} & 
\cdots & 
\frac{\partial f_1}{\partial x_n} \\[1ex] 
\vdots & 
\ddots  & 
\vdots \\[1ex]
\frac{\partial f_m}{\partial x_1} & 
\cdots & 
\frac{\partial f_m}{\partial x_n}
\end{bmatrix},
\end{equation}
and whose $(i,j)$-th entry is $\mathbf{J}_{ij}=\frac{\partial f_i}{\partial x_j}$. The expression $\text{vec}\{\mathbf{X}\}$ coverts $f(\bm{x})$, which potentially results in higher dimensional representations, into a vector representation.
Along with the chain rule of differentiation, the Jacobian matrix can then be used to calculate the gradients of the loss function with respect to the weights and biases of that layer. The chain rule is used to differentiate composite function and it states, that if $y = f(\bm{u})$, where $\bm{u} = g(\bm{x})$, the following holds
\begin{equation}
\frac{\partial (\text{vec}\{\bm{y}\})}{\partial  (\text{vec}\{\bm{x}\})^\textrm{T}} = \frac{\partial (\text{vec}\{\bm{y}\})}{\partial  (\text{vec}\{\bm{u}\})^\textrm{T}} \cdot \frac{\partial (\text{vec}\{\bm{u}\})}{\partial  (\text{vec}\{\bm{x}\})^\textrm{T}}.
\end{equation}
Through the repeated usage of the chain rule, the gradients of the networks' parameters can be determined.
\section{Gradients of GATv2 network parameters}
In this section, the derivatives for the trainable parameters of the GATv2\cite{GATv2} architecture are derived. The gradients for updating the weights and biases due to node $i$ also depend on its neigboring nodes $j\in \mathcal{N}_i$. In total, node $i$ has $N = |\mathcal{N}_i|$ neighboring nodes.
The trainable network parameters of a one-layered GATv2 are $\mathbf{\Theta}_R, \mathbf{\Theta}_L \in \mathbb{R}^{D\times (H+1)}$, $\bm{a} \in \mathbb{R}^{D}$ and $\bm{b}\in\mathbb{R}^{D}$. Note, that the parameter \mbox{$\boldsymbol{\Theta}_{R}= [\boldsymbol{\Theta}_{R_b}, \boldsymbol{\Theta}_{R_w}]$} (as well as $\mathbf{\Theta}_L$) is composed of a weight term ~$\boldsymbol{\Theta}_{R_w}$ and a bias term~$\boldsymbol{\Theta}_{R_b}$.
The considered loss function $\mathcal{L}(\bm{h}_{i}', \bm{y})$ does not include any intermediate loss or regularization terms but is based only on the node-level prediction error. In Fig.~\ref{fig:fbpass}, the partial forward and backward pass of a one-layered GATv2 network for updating the parameter $\mathbf{\Theta}_R$ is shown. In the following, the expression $\text{vec}\{\mathbf{X}\}$ indicates that $\mathbf{X}$ is represented through vector formulation and superscripts indicate the entry index of a vector representation. The parameter $c$ indicates the slope for the negative value range in the LeakyReLU activation function.
\input{forwardbackwardpass.tex} 
\subsection{Gradient for parameter $\mathbf{\Theta}_R$}
\begingroup
\allowdisplaybreaks
\begin{align*}
&\mathbf{J}_{a_{9,i1}} = \frac{\partial (\text{vec}\{\bm{h}_{i}'\})}{\partial (\text{vec}\{\bm{a}_{9,i1}\})^\mathrm{T}} =
\frac{\partial
(\text{vec}\{\sum_{j \in \mathcal{N}_i} \bm{a}_{9,ij}\})}{\partial (\text{vec}\{\bm{a}_{9,i1}\})^\mathrm{T}} = 
\begin{bmatrix}
\frac{\partial (\text{vec}\{\sum_{j \in \mathcal{N}_i} a_{9,ij}\}^{(1)})}{\partial a_{9,i1}^{(1)}} & \cdots & \frac{\partial( \text{vec}\{\sum_{j \in \mathcal{N}_i} a_{9,ij}\}^{(1)})}{\partial a_{9,i1}^{(D)}} \\
\vdots & \ddots & \vdots\\
\frac{\partial (\text{vec}\{\sum_{j \in \mathcal{N}_i} a_{9,ij}\}^{(D)})}{\partial a_{9,i1}^{(1)}} & \cdots & \frac{\partial(\text{vec}\{\sum_{j \in \mathcal{N}_i} a_{9,ij}\}^{(D)})}{\partial a_{9,i1}^{(D)}} \\
\end{bmatrix}\\
&\hspace{0.9cm}=
\begin{bmatrix}
1 & 0 & \cdots & 0 \\
0 & 1 & \cdots & \vdots \\
\vdots &\ddots & \ddots & 0 \\
0 & \cdots & 0& 1
\end{bmatrix} = \mathbf{I}\\
&\mathbf{J}_{a_{8,i1}}  = \mathbf{J}_{a_{9,i1}} \cdot
\frac{\partial (\text{vec}\{\bm{a}_{9,i1}\})}{\partial (\text{vec}\{\bm{a}_{8,i1}\})^\mathrm{T}} = 
\mathbf{J}_{a_{9,i1}} \cdot
\frac{\partial (\text{vec}\{({\bm{a}}_{8,i1}\cdot\tilde{\bm{h}}_1))\}}{\partial (\text{vec}\{\bm{a}_{8,i1}\})^\mathrm{T}} = \mathbf{I} \cdot
\begin{bmatrix}
\tilde{h}_1^\textrm{T} & \bm{0}_{1 \times H} & \cdots & \bm{0}_{1 \times H}  \\
\bm{0}_{1 \times H} & \tilde{h}_1^\textrm{T}  & \cdots & \vdots \\
\vdots &\ddots & \ddots & \bm{0}_{1 \times H} \\
\bm{0}_{1 \times H} & \cdots & \bm{0}_{1 \times H}& \tilde{h}_1^\textrm{T}
\end{bmatrix} \\
&\hspace{0.9cm}=
\begin{bmatrix}
\tilde{h}_1^\textrm{T} & \bm{0}_{1 \times H} & \cdots & \bm{0}_{1 \times H}  \\
\bm{0}_{1 \times H} & \tilde{h}_1^\textrm{T}  & \cdots & \vdots \\
\vdots &\ddots & \ddots & \bm{0}_{1 \times H} \\
\bm{0}_{1 \times H} & \cdots & \bm{0}_{1 \times H}& \tilde{h}_1^\textrm{T}
\end{bmatrix}\\
\vspace{1cm}\\
&{J}_{a_{7,i1}}  =  \mathbf{J}_{a_{8,i1}}  \cdot
\frac{\partial (\text{vec}\{\bm{a}_{8,i1}\})}{\partial (\text{vec}\{{a}_{7,i1}\})^\textrm{T}} 
= 
\mathbf{J}_{a_{8,i1}}  \cdot
\frac{\partial (\text{vec}\{(\mathbf{\Theta}_L\cdot{a}_{7,i1})\})}{\partial ({a}_{7,i1})^\textrm{T}} 
= \mathbf{J}_{a_{8,i1}} \cdot \text{vec}\{(\mathbf{\Theta}_L)\} \\
&\hspace{0.9cm}=
\begin{bmatrix}
\tilde{\bm{h}}_1^\textrm{T} & \bm{0}_{1 \times H} & \cdots & \bm{0}_{1 \times H}  \\
\bm{0}_{1 \times H} & \tilde{\bm{h}}_1^\textrm{T}  & \cdots & \vdots \\
\vdots &\ddots & \ddots & \bm{0}_{1 \times H} \\
\bm{0}_{1 \times H} & \cdots & \bm{0}_{1 \times H}& \tilde{\bm{h}}_1^\textrm{T}
\end{bmatrix} \cdot \text{vec}\{(\mathbf{\Theta}_L)\} =
\begin{bmatrix}
\tilde{\bm{h}}_1^\textrm{T} \mathbf{\Theta}_{L}^{(1)} \\
\tilde{\bm{h}}_1^\textrm{T} \mathbf{\Theta}_{L}^{(2)} \\
\vdots\\
\tilde{\bm{h}}_1^\textrm{T} \mathbf{\Theta}_{L}^{(D)} \\
\end{bmatrix} = \mathbf{\Theta}_L \tilde{\bm{h}}_1 \rightarrow \sum_{d=1}^D (\mathbf{\Theta}_L \tilde{\bm{h}}_1)^{(d)} \\ 
\vspace{1cm}\\
&\mathbf{J}_{a_{7,i}}  = concat_j ( {J}_{a_{7,ij}}  ) = 
\left[ {J}_{a_{7,i1}}\quad {J}_{a_{7,i2}} \quad \cdots\quad {J}_{a_{7,iN}} \right]=\begin{bmatrix}
A_{7,i1} & A_{7,i2} & \cdots & A_{7,iN}\\
\end{bmatrix}\\
&\mathbf{J}_{a_{6,i}}=  \mathbf{J}_{a_{7,i}} \cdot
\frac{\partial (\text{vec}\{\bm{a}_{7,i}\})}{\partial (\text{vec}\{\bm{a}_{6,i}\})^\textrm{T}} 
= \mathbf{J}_{a_{7,i}} \cdot
\begin{bmatrix}
(1-\alpha_{i1}) & -\alpha_{i1}\alpha_{i2} &-\alpha_{i1}\alpha_{i3} & \cdots & -\alpha_{i1}\alpha_{iN}\\
-\alpha_{i1}\alpha_{i2}  & \alpha_{i2}(1-\alpha_{i2}) & -\alpha_{i2}\alpha_{i3} & \cdots & \vdots\\
-\alpha_{i1}\alpha_{i3}  & -\alpha_{i2}\alpha_{i3} & \alpha_{i3}(1-\alpha_{i3}) & \cdots & \vdots\\
\vdots  &  \ddots  & \ddots & \ddots & \vdots\\
-\alpha_{i1}\alpha_{iN}  &  -\alpha_{i2}\alpha_{iN}  & \cdots & \cdots & \alpha_{iN}(1-\alpha_{iN})\\
\end{bmatrix}\\
&\hspace{0.9cm}=
\begin{bmatrix}
\sum_j^N \alpha_{i1} (\delta_{1j} -\alpha_{ij})A_{7,ij}\\
\sum_j^N \alpha_{i2} (\delta_{2j} -\alpha_{ij})A_{7,ij} \\
\vdots\\
\sum_j^N \alpha_{iN} (\delta_{Nj} -\alpha_{ij})A_{7,ij} \\
\end{bmatrix}^{\textrm{T}}, \quad \text{ where } \delta_{lj} =
\begin{cases}
1, & \text{if}\ l=j\\
0, & \text{else}
\end{cases}\\
\vspace{1cm}\\
&\mathbf{J}_{a_{6,i1}} = 
\mathbf{J}_{a_{6,i}} \cdot
\frac{\partial( \text{vec}\{\bm{a}_{6,i}\})}{\partial (\text{vec}\{\bm{a}_{6,i1}\})^\textrm{T}}  = \mathbf{J}_{a_{6,i1}} \cdot \left[1\quad 0 \quad \cdots \quad 0\right]^\textrm{T} \\
&\hspace{0.9cm}=
\begin{bmatrix}
\sum_j^N \alpha_{i1} (\delta_{1j} -\alpha_{ij}) A_{7,ij}
\end{bmatrix}, \quad \text{ where } \delta_{1j} =
\begin{cases}
1, & \text{if}\ j=1\\
0, & \text{else}
\end{cases}\\
\vspace{1cm}\\
&\mathbf{J}_{a_{5,i1}} = \mathbf{J}_{a_{6,i1}} \cdot
\frac{\partial (\text{vec}\{\bm{a}_{6,i1}\})}{\partial (\text{vec}\{\bm{a}_{5,i1}\})^\textrm{T}} 
= \mathbf{J}_{a_{6,i1}} \cdot \frac{\partial (\text{vec}\{\bm{a}_{5,i1} \odot \bm{a}\})}{\partial (\text{vec}\{\bm{a}_{5,i1}\})^\textrm{T}} =
\mathbf{J}_{a_{6,i1}} \cdot 
\begin{bmatrix}
{a}^{(1)}  &
{a}^{(2)}  & \cdots & {a}^{(D)}\\
\end{bmatrix}\\ 
&\hspace{0.9cm}=
\begin{bmatrix}
\sum_j^N \alpha_{i1} (\delta_{1j} -\alpha_{ij}) A_{7,ij}
\end{bmatrix}
\cdot
\begin{bmatrix}
{a}^{(1)}  &
{a}^{(2)}  & \cdots & {a}^{(D)}\\
\end{bmatrix}\\
&\hspace{0.9cm}=
\begin{bmatrix}
a^{(1)} \sum_j^N \alpha_{i1} (\delta_{1j} -\alpha_{ij}) A_{7,ij}\\
a^{(2)} \sum_j^N \alpha_{i1} (\delta_{1j} -\alpha_{ij}) A_{7,ij} \\
\vdots\\
a^{(D)} \sum_j^N \alpha_{i1} (\delta_{1j} -\alpha_{ij}) A_{7,ij} \\
\end{bmatrix}^{\textrm{T}}\\
&\mathbf{J}_{a_{4,i1}} = \mathbf{J}_{a_{5,i1}}\cdot
\frac{\partial (\text{vec}\{\bm{a}_{5,i1}\})}{\partial (\text{vec}\{\bm{a}_{4,i1}\})^\textrm{T}}
= \mathbf{J}_{a_{5,i1}} \cdot \frac{\partial (\text{LeakyRELU}(\text{vec}\{\bm{a}_{5,i1}\}))}{\partial (\text{vec}\{\bm{a}_{4,i1}\})^\textrm{T}}
= \mathbf{J}_{a_{5,i1}}  \cdot 
\begin{bmatrix}
s_{i1}^{(1)} & 0 & \cdots & 0 \\
0 & s_{i1}^{(2)} & \cdots & 0  \\
\vdots & \ddots & \ddots & 0  \\
0 & 0 & \cdots & s_{i1}^{(D)}  \\
\end{bmatrix}\\
&\hspace{0.9cm}=
\begin{bmatrix}
s_{i1}^{(1)}a^{(1)} \sum_j^N(\alpha_{i1} (\delta_{1j} -\alpha_{ij}) A_{7,ij})\\
s_{i1}^{(2)}a^{(2)} \sum_j^N(\alpha_{i1} (\delta_{1j} -\alpha_{ij}) A_{7,ij})\\
\vdots \\
s_{i1}^{(D)}a^{(D)} \sum_j^N(\alpha_{i1} (\delta_{1j} -\alpha_{ij}) A_{7,ij})\\
\end{bmatrix}^\textrm{T},\quad \text{ where } s_{i1}^{(d)} =     
\begin{cases}
1, & \text{if}\ a_{4,i1}^{(d)}>0 \\
s_n, & \text{else}
\end{cases}\\
\vspace{1cm}\\
&\mathbf{J}_{a_{2,i1}} = \mathbf{J}_{a_{4,i1}}\cdot
\frac{\partial (\text{vec}\{\bm{a}_{4,i1}\})}{\partial (\text{vec}\{\bm{a}_{2,i1}\})^\textrm{T}}
=\mathbf{J}_{a_{4,i1}} \cdot
\mathbf{I} = \mathbf{J}_{a_{4,i1}}\\
& {}^\text{B}\mathbf{\Theta}_{R_w,i1}^{\textrm{T}} 
= \frac{\partial \bm{h}_i'}{\partial \mathbf{\Theta}_{R_w, i1}} 
= \frac{\partial (\text{vec}\{\bm{h}_{i}'\})}{\partial (\text{vec}\{\bm{a}_{2,i1}\})^\textrm{T}}\cdot
\frac{\partial (\text{vec}\{\bm{a}_{2,i1}\})}{\partial (\text{vec}\{\mathbf{\Theta}_{R}\})^\textrm{T}} \\
&\hspace{0.9cm}
= \mathbf{J}_{a_{2,i1}} \cdot \frac{\partial (\text{vec}\{\mathbf{\Theta}_{R} h_i\})}{\partial (\text{vec}\{\mathbf{\Theta}_{R}\})^\textrm{T}}
= \mathbf{J}_{a_{2,i1}} \cdot 
\begin{bmatrix} \!
\tilde{\bm{h}}_1^\textrm{T} & \bm{0}_{1 \times H} & \cdots & \bm{0}_{1 \times H}  \\
\bm{0}_{1 \times H} & \tilde{\bm{h}}_1^\textrm{T}  & \cdots & \vdots \\
\vdots &\ddots & \ddots & \bm{0}_{1 \times H} \\
\bm{0}_{1 \times H} & \cdots & \bm{0}_{1 \times H}& \tilde{\bm{h}}_1^\textrm{T}
\end{bmatrix} =
\begin{bmatrix}
h_i^{(1)} s_{i1}^{(1)}a^{(1)} \sum_j^N(\alpha_{i1} (\delta_{1j} -\alpha_{ij}) A_{7,ij})\\
h_i^{(2)} s_{i1}^{(1)}a^{(2)} \sum_j^N(\alpha_{i1} (\delta_{1j} -\alpha_{ij}) A_{7,ij})\\
\vdots\\
h_i^{(D)} s_{i1}^{(1)}a^{(2)} \sum_j^N(\alpha_{i1} (\delta_{1j} -\alpha_{ij}) A_{7,ij})\\
h_i^{(1)} s_{i1}^{(2)}a^{(2)} \sum_j^N(\alpha_{i1} (\delta_{1j} -\alpha_{ij}) A_{7,ij})\\
\vdots\\
h_i^{(D)} s_{i1}^{(D)}a^{(2)} \sum_j^N(\alpha_{i1} (\delta_{1j} -\alpha_{ij}) A_{7,ij})\\
\end{bmatrix}^\textrm{T} \\
&\hspace{0.9cm}=
\begin{bmatrix}
h_i^{(1)} s_{i1}^{(1)}a^{(1)}\alpha_{i1} \sum_j^N((\delta_{1j} -\alpha_{ij}) A_{7,ij})\\
h_i^{(2)} s_{i1}^{(1)}a^{(2)}\alpha_{i1} \sum_j^N((\delta_{1j} -\alpha_{ij}) A_{7,ij})\\
\vdots\\
h_i^{(D)} s_{i1}^{(1)}a^{(2)}\alpha_{i1} \sum_j^N((\delta_{1j} -\alpha_{ij}) A_{7,ij})\\
h_i^{(1)} s_{i1}^{(2)}a^{(2)}\alpha_{i1} \sum_j^N((\delta_{1j} -\alpha_{ij}) A_{7,ij})\\
\vdots\\
h_i^{(D)} s_{i1}^{(D)}a^{(2)}\alpha_{i1} \sum_j^N((\delta_{1j} -\alpha_{ij}) A_{7,ij})\\
\end{bmatrix}^\textrm{T}
=
\begin{bmatrix}
h_i^{(1)} s_{i1}^{(1)}a^{(1)} \alpha_{i1} S_1\\
h_i^{(2)} s_{i1}^{(1)}a^{(2)} \alpha_{i1} S_1\\
\vdots \\
h_i^{(D)} s_{i1}^{(1)}a^{(2)}\alpha_{i1} S_1\\
h_i^{(1)} s_{i1}^{(2)}a^{(2)}\alpha_{i1} S_1\\
\vdots\\
h_i^{(D)} s_{i1}^{(D)}a^{(2)}\alpha_{i1} S_1\\
\end{bmatrix}^\textrm{T}\\
&\hspace{0.9cm}\\
&{}^\text{B}\mathbf{\Theta}_{R_w,i}^{\textrm{T}}=  \frac{\partial \bm{h}_i'}{\partial \mathbf{\Theta}_{R_w, i}} =  \sum_k^N {}^\text{B}\mathbf{\Theta}_{R_w,ik}^{\textrm{T}} = \sum_k^N
\begin{bmatrix}
h_i^{(1)} s_{ik}^{(1)}a^{(1)}\alpha_{ik} S_k\\
\vdots \\
h_i^{(D)} s_{ik}^{(D)}a^{(D)}\alpha_{ik} S_k\\
\end{bmatrix}^\textrm{T}\\
\vspace{1cm}\\
&\text{Hence, the gradient of $\bm{h}_i'$ with regard to the $t$-th entry of the vectorized parameter $\mathbf{\Theta}_{R_{w,i}}$ is }\\
&{}^\text{B}{\Theta}_{R_w,i}^{(t)}= \sum_k^N {}^\text{B}{\Theta}_{R_w,ik}^{(t)}  =
\sum_k^N 
\begin{bmatrix}
h_i^{(t)}a^{(t)} s_{ik}^{(t)}\alpha_{ik} S_k
\end{bmatrix} = h_i^{(t)}a^{(t)} \sum_k^N
\begin{bmatrix}
s_{ik}^{(t)}\alpha_{ik} S_k
\end{bmatrix} \\
& \hspace{0.9cm} \text{ where } S_k = \sum_j^N (\delta_{kj} -\alpha_{ij})A_{7,ij}, \quad \text{ with } \delta_{kj} =
\begin{cases}
1, & \text{if}\ k=j\\
0, & \text{else}
\end{cases}\\
& \hspace{0.9cm} \text{ and } A_{7,ij} = \sum_d^D \left(\mathbf{\Theta}_L \tilde{\bm{h}}_j\right)^{(d)}. \\
&\text{The mathematical expression can be reformulated as follows.}\\
&{}^\text{B}{\Theta}_{R_w,i}^{(t)}= h_i^{(t)}a^{(t)} \sum_k^N 
\begin{bmatrix}
s_{ik}^{(t)}\alpha_{ik} S_k
\end{bmatrix} = h_i^{(t)}a^{(t)} \sum_k^N 
\begin{bmatrix}
s_{ik}^{(t)}\alpha_{ik} \sum\limits_{j}^N  (\delta_{kj} -\alpha_{ij})A_{7,ij}
\end{bmatrix} \\ 
& \hspace{0.9cm} = h_i^{(t)}a^{(t)} \sum_k^N 
\begin{bmatrix}
s_{ik}^{(t)}\alpha_{ik} \left(\sum\limits_{j\neq k}^N  ( -\alpha_{ij})A_{7,ij} + (1 -\alpha_{ik})A_{7,ik} \right)
\end{bmatrix}, \quad \text{ where } \alpha_{ik} = (1-\sum\limits_{j\neq k}^N  \alpha_{ij})\\
& \hspace{0.9cm} = h_i^{(t)}a^{(t)} \sum_k^N  
\begin{bmatrix}
s_{ik}^{(t)}\alpha_{ik} \left(\sum\limits_{j\neq k}^N   ( -\alpha_{ij})A_{7,ij} + (1 - (1-\sum_{j\neq k}^N \alpha_{ij}))A_{7,ik} \right)
\end{bmatrix}\\
& \hspace{0.9cm} = h_i^{(t)}a^{(t)} \sum_k^N  
\begin{bmatrix}
s_{ik}^{(t)}\alpha_{ik} \left(\sum\limits_{j\neq k}^N   ( -\alpha_{ij})A_{7,ij} + (\sum\limits_{j\neq k}^N \alpha_{ij})A_{7,ik} \right)
\end{bmatrix}\\
& \hspace{0.9cm} = h_i^{(t)}a^{(t)} \sum_k^N  
\begin{bmatrix}
s_{ik}^{(t)}  \left(\sum\limits_{j\neq k}^N   \alpha_{ik} \alpha_{ij}  (A_{7,ik}-A_{7,ij})  \right)
\end{bmatrix}\\
& \hspace{0.9cm} = h_i^{(t)}a^{(t)} \sum_k^N  \sum\limits_{j\neq k}^N  
\begin{bmatrix}
s_{ik}^{(t)}  \alpha_{ik} \alpha_{ij}  (A_{7,ik}-A_{7,ij})  
\end{bmatrix}\\
& \hspace{0.9cm} = h_i^{(t)}a^{(t)} \sum_k^N  \sum\limits_{j\neq k}^N \frac{1}{2}
\begin{bmatrix} 2\,
s_{ik}^{(t)}  \alpha_{ik} \alpha_{ij}  (A_{7,ik}-A_{7,ij})  
\end{bmatrix}\\
& \hspace{0.9cm} = h_i^{(t)}a^{(t)} \sum_k^N  \sum_{j\neq k}^N \frac{1}{2}
\begin{bmatrix} 
s_{ik}^{(t)}  \alpha_{ik} \alpha_{ij}  (A_{7,ik}-A_{7,ij}) +
s_{ik}^{(t)}  \alpha_{ik} \alpha_{ij}  (A_{7,ik}-A_{7,ij})   
\end{bmatrix}\\
& \hspace{0.9cm} = h_i^{(t)}a^{(t)} \left( \sum_k^N  \sum_{j\neq k}^N \frac{1}{2}
\begin{bmatrix} 
s_{ik}^{(t)}  \alpha_{ik} \alpha_{ij}  (A_{7,ik}-A_{7,ij})\end{bmatrix} + \sum_k^N  \sum_{j\neq k}^N \frac{1}{2}
\begin{bmatrix}  
s_{ik}^{(t)}  \alpha_{ik} \alpha_{ij}  (A_{7,ik}-A_{7,ij})   
\end{bmatrix}\right)\\
& \hspace{0.9cm} = h_i^{(t)}a^{(t)} \left( \sum_k^N  \sum_{j\neq k}^N \frac{1}{2}
\begin{bmatrix} 
s_{ik}^{(t)}  \alpha_{ik} \alpha_{ij}  (A_{7,ik}-A_{7,ij})\end{bmatrix} - \sum_k^N  \sum_{j\neq k}^N \frac{1}{2}
\begin{bmatrix}  
s_{ik}^{(t)}  \alpha_{ik} \alpha_{ij}  (A_{7,ij}-A_{7,ik})   
\end{bmatrix}\right)\\
& \hspace{0.9cm} = h_i^{(t)}a^{(t)} \left( \sum_k^N  \sum_{j\neq k}^N \frac{1}{2}
\begin{bmatrix} 
s_{ik}^{(t)}  \alpha_{ik} \alpha_{ij}  (A_{7,ik}-A_{7,ij})\end{bmatrix} - \sum_j^N  \sum_{k\neq j}^N \frac{1}{2}
\begin{bmatrix}  
s_{ij}^{(t)}  \alpha_{ik} \alpha_{ij}  (A_{7,ik}-A_{7,ij})   
\end{bmatrix}\right)\\
& \hspace{0.9cm} =h_i^{(t)}a^{(t)}  \sum_{k}^{N} \sum_{j}^{N} \frac{1}{2}
\begin{bmatrix}
\alpha_{ik} \alpha_{ij}  (A_{7,ik}-A_{7,ij}) (s_{ik}^{(t)}- s_{ij}^{(t)})
\end{bmatrix}, \quad j\neq k \\
& \hspace{0.9cm} =h_i^{(t)}a^{(t)} \sum_{j,k}^{\mathcal{S}_i}
\begin{bmatrix}
\alpha_{ik} \alpha_{ij}  (A_{7,ik}-A_{7,ij}) (s_{ik}^{(t)}- s_{ij}^{(t)})
\end{bmatrix}, \\
& \hspace{0.9cm} \text{ where $\mathcal{S}_i=[\mathcal{N}_i]^2$ is the set of all the subsets of $\mathcal{N}_i$ with exactly two elements and no pairing repetitions.}\\
&\text{With regard to the loss function $\mathcal{L}$, the gradient of the matrix elements ${\Theta}_{R_{w,i}}^{(t)}$ and ${\Theta}_{R_{w,i}}^{(t)}$ are }\\
&{}^\text{B}\Theta_{R_w, i}^{(t)} = \frac{\partial \mathcal{L}}{\partial {h}_i'^{(t)}}\!\cdot\!
\left[
h_i^{(t)} a^{(t)}\sum_{j,k}^{\mathcal{S}_i} \alpha_{ij}\alpha_{ik} (A_j\! - \!A_k)(s_{ij}^{(t)}-s_{ik}^{(t)})
\right]\!, \quad \text{and}\\
&{}^\text{B}\Theta_{R_b, i}^{(t)} = \frac{\partial \mathcal{L}}{\partial {h}_i'^{(t)}}\!\cdot\!
\left[
a^{(t)}\sum_{j,k}^{\mathcal{S}_i} \alpha_{ij}\alpha_{ik} (A_j\! - \!A_k)(s_{ij}^{(t)}-s_{ik}^{(t)})
\right]\!,\\
& \hspace{0.9cm} \text{ where } s_{i\{j,k\}}^{(t)} =     
\begin{cases}
1, & \text{if}\ a_{4,i\{j,k\}}^{(t)}>0 \\
s_n, & \text{else}
\end{cases},\\
& \hspace{0.9cm} \text{ and } A_{7,ij} = \sum_d^D \left(\mathbf{\Theta}_L \tilde{\bm{h}}_j\right)^{(d)}.
\end{align*}
\endgroup
\subsection{Gradient for parameter $\mathbf{\Theta}_L$}
With regard to the loss function $\mathcal{L}$, the gradient of the matrix elements ${\Theta}_{L_{w,i}}^{(t)}$ and ${\Theta}_{L_{w,i}}^{(t)}$ are
\begin{flalign*}
&{}^\text{B}\Theta_{L_w, i}^{(t)} = \frac{\partial \mathcal{L}}{\partial {h}_i'^{(t)}}\!\cdot\!
\sum_k^N 
\begin{bmatrix}
h_k^{(t)}a^{(t)} s_{ik}^{(t)}\alpha_{ik} S_k +  
\alpha_{ik} h_k^{(t)}
\end{bmatrix},\quad \text{and}\quad\quad\quad\quad\quad \quad\quad\quad\quad\quad\quad\quad\quad\quad\quad\quad\quad\quad\quad\quad\quad\quad\quad\quad\quad\quad\text{   }\\
&{}^\text{B}\Theta_{L_b, i}^{(t)} = \frac{\partial \mathcal{L}}{\partial {h}_i'^{(t)}}\!\cdot\!
\sum_k^N 
\begin{bmatrix}
a^{(t)} s_{ik}^{(t)}\alpha_{ik} S_k +  
\alpha_{ik} 
\end{bmatrix}.
\end{flalign*}
\subsection{Gradient for parameter $\bm{b}$}
With regard to the loss function $\mathcal{L}$, the gradient of the bias $\bm{b}$ is
\begin{flalign*}
&{}^\text{B}\bm{b}^{\mathrm{T}} = \frac{\partial \mathcal{L}}{\partial \bm{h}_i'}\cdot \frac{\partial (\text{vec}\{\bm{\tilde{h}}_i'\})}{\partial (\text{vec}\{\bm{b}\})^\mathrm{T}} 
=\frac{\partial \mathcal{L}}{\partial \bm{h}_i'}\cdot \frac{\partial (\text{vec}\{\sum_{j \in \mathcal{N}_i} a_{9,ij} +\bm{b}\})}{\partial (\text{vec}\{\bm{b}\})^\mathrm{T}}= \frac{\partial \mathcal{L}}{\partial \bm{h}_i'}\cdot
\begin{bmatrix}
1 & 0 & \cdots & 0 \\
0 & 1 & \cdots & \vdots \\
\vdots &\ddots & \ddots & 0 \\
0 & \cdots & 0& 1
\end{bmatrix} = \frac{\partial \mathcal{L}}{\partial \bm{h}_i'}. \quad \quad \quad\quad \quad \quad \quad \quad\quad\, \,  \, \,  \text{  }\\
\end{flalign*}
\newpage
 {
 	\bibliographystyle{IEEEtranS}
 	\bibliography{ref_backprop.bib}
 }
\end{document}

%% file: forwardbackwardpass.tex
\tikzset{rechteck/.style={rectangle, minimum width=10mm, minimum height=8mm, draw=black, fill=white, align=center}}

\tikzset{Kreis/.style={circle,draw=black, fill=white!10, minimum size=1cm, inner sep=0pt, outer sep=0pt}}
\tikzset{brechteck/.style={rectangle, minimum width=10mm, minimum height=8mm, draw=blue,  text=blue, fill=white, align=center}}

\tikzset{bKreis/.style={circle,draw=blue, text=blue, fill=white!10, minimum size=1cm, inner sep=0pt, outer sep=0pt}}
\begin{figure}[t!]
	\centering
	\scalebox{.63}{ 
		\begin{tikzpicture}
		\def \d {1.7} 
		\def \h {1.2}
		\node (a0)[Kreis] at (0, 1) {$\tilde{\bm{h}}_i$}; 
		\node (a1)[Kreis] at (0,-1) {$\tilde{\bm{h}}_1$};
		
		\node (g0)[rechteck] at (\d,1) {$g_0$};
		\node (g1)[rechteck] at (\d,-1) {$g_1$};
		\node (tR)[Kreis] at ($(g0) + (0, \h)$) {$\mathbf{\Theta}_R$};
		\node (tL)[Kreis] at ($(g1) + (0, -\h)$) {$\mathbf{\Theta}_L$};
		
		\node (a2)[Kreis] at ($(g0) + (\d,0)$) {$\bm{a}_{2,i1}$};
		\node (a3)[Kreis] at ($(g1) + (\d,0)$) {$\bm{a}_{3,i1}$};
		
		\node (g2)[rechteck] at ($(a2)!0.5!(a3) + (\d,0)$) {$g_2$\\$\bm{a}_{2,i1}+\bm{a}_{3,i1}$};
		\node (a4)[Kreis] at ($(g2) + (\d,0)$) {$\bm{a}_{4,i1}$};
		\node (g3)[rechteck] at ($(a4) + (\d,0)$) {$g_3$\\$LR(a_4)$};
		\node (a5)[Kreis] at ($(g3) + (\d,0)$) {$\bm{a}_{5,i1}$};
		\node (g4)[rechteck] at ($(a5) + (\d,0)$) {$g_4$\\$(\bm{a}_{5,i1}\odot\bm{a})$};
		
		\node (aT)[Kreis] at ($(g4) + (0,\h)$) {$\bm{a}$};
		
		\node (a61)[Kreis] at ($(g4) + (\d,0)$) {$a_{6,i1}$};
		\node (a62)[Kreis] at ($(g4) + (\d,-2*\h)$) {$a_{6,iN}$};
		\path (a61) -- (a62) node [font=\Huge, midway, sloped] {$\dots$};
		
		\node (g5)[rechteck] at ($(a61)!0.5!(a62) + (\d,0)$) {$g_{5}$\\sftmx()};
		
		\node (a71)[Kreis] at ($(a61) + (2*\d,0)$) {$a_{7,i1}$};
		\node (a72)[Kreis] at ($(a62) + (2*\d,0)$) {$a_{7,iN}$};
		\path (a71) -- (a72) node [font=\Huge, midway, sloped] {$\dots$};		
		\node (g6)[rechteck] at ($(a71) + (\d,0)$) {$g_6$};
		\node (g7)[rechteck] at ($(a72) + (\d,0)$) {$g_7$};

		\node (tN)[Kreis] at ($(g6) + (0,\h)$) {$\mathbf{\Theta}_L$};
		\node (tLL)[Kreis] at ($(g7) + (0,-\h)$) {$\mathbf{\Theta}_L$};
		
		\node (a81)[Kreis] at ($(g6) + (\d,0)$) {$\bm{a}_{8,i1}$};
		\node (a82)[Kreis] at ($(g7) + (\d,0)$) {$\bm{a}_{8,iN}$};
		
		\node (g8)[rechteck] at ($(a81) + (\d,0)$) {$g_8$};
		\node (g9)[rechteck] at ($(a82) + (\d,0)$) {$g_9$};	
		
		\node (hii)[Kreis] at ($(g8) + (0 , \h)$) {$\tilde{\bm{h}}_1$};
		\node (hjj)[Kreis] at ($(g9) + (0 , -\h)$) {$\tilde{\bm{h}}_N$};
		
		\node (a9)[Kreis] at ($(g8) + (\d,0)$) {$\bm{a}_{9,i1}$};
		\node (a10)[Kreis] at ($(g9) + (\d,0)$) {$\bm{a}_{9,iN}$};
		
		\node (g10)[rechteck] at ($(a9)!0.5!(a10) + (\d,0)$) {$g_{10}$\\$\sum_{j} \bm{a}_{9,ij}+ \bm{b}$};
		\node (b)[Kreis] at ($(g10) + (0, \h)$) {$\bm{b}$};
		\node (a11)[Kreis] at ($(g10) + (\d,0)$) {$\bm{h}_{i}'$};
		
		\draw[->] (a0)  -- (g0);
		\draw[->] (a1)  -- (g1);
		\draw[->] (g0)  -- (a2);
		\draw[->] (g1)  -- (a3);
		
		\draw[->] (tR)  -- (g0);
		\draw[->] (tL)  -- (g1);
		
		\draw[->] (a2)  -- (g2);
		\draw[->] (a3)  -- (g2);
		
		\draw[->] (g2)  -- (a4);
		\draw[->] (a4)  -- (g3);
		\draw[->] (g3)  -- (a5);
		\draw[->] (a5)  -- (g4);
		\draw[->] (aT)  -- (g4);
		\draw[->] (g4)  -- (a61);
		\draw[->] (a61)  -- (g5);
		\draw[->] (a62)  -- (g5);
		\draw[->] (g5)  -- (a71);
		\draw[->] (g5)  -- (a72);
		\draw[->] (a71)  -- (g6);
		\draw[->] (a72)  -- (g7);
		\draw[->] (tN)  -- (g6);
		\draw[->] (tLL)  -- (g7);
		\draw[->] (g6)  -- (a81);
		\draw[->] (g7)  -- (a82);
		\draw[->] (a81)  -- (g8);
		\draw[->] (a82)  -- (g9);
		\draw[->] (hii)  -- (g8);
		\draw[->] (hjj)  -- (g9);
		\draw[->] (g8)  -- (a9);
		\draw[->] (g9)  -- (a10);
		\draw[->] (a9)  -- (g10);
		\draw[->] (a10)  -- (g10);
		\draw[->] (a81)  -- (g8);
		\draw[->] (g10)  -- (a11);
		\draw[->] (b) -- (g10);
		\node(ba11)[bKreis] at ($(a11) -(0, 5)$) {${}^\text{B}\bm{h}_{i}^\textrm{'T}$};
				\path (a11) -- (ba11) node(L) [font=\huge, midway] {$\mathcal{L}$};
				\draw[->] (a11)  -- (L);
				\draw[->,blue] (L)  -- (ba11);
		\node(bg10)[brechteck] at ($(ba11) -(\d, 0)$) {${}^\text{B}g_{10}^\textrm{T}$};
		\node(ba9)[bKreis] at ($(bg10) -(\d,0)$) {${}^\text{B}\bm{a}_{9,i1}^\textrm{T}$};
		\node(ba10)[bKreis] at ($(ba9) -(0,1.5*\h)$) {${}^\text{B}\bm{a}_{9,iN}^\textrm{T}$};
		\node(bg8)[brechteck] at ($(ba9) -(\d, 0)$) {${}^\text{B}g_{8}^\textrm{T}$};
		\node(ba81)[bKreis] at ($(bg8) -(\d,0)$) {${}^\text{B}\bm{a}_{8,i1}^\textrm{T}$};
		\node(bg6)[brechteck] at ($(ba81) -(\d, 0)$) {${}^\text{B}g_{6}^\textrm{T}$};
		\node(ba71)[bKreis] at ($(bg6) -(\d,0)$) {${}^\text{B}\bm{a}_{7,i1}^\textrm{T}$};
		\node(ba72)[bKreis] at ($(ba71) + (0,-1.5*\h)$) {${}^\text{B}\bm{a}_{7,iN}^\textrm{T}$};
		\node(bg5)[brechteck] at ($(ba71) -(\d, 0)$) {${}^\text{B}g_{5}^\textrm{T}$};
		\node(ba61)[bKreis] at ($(bg5) -(\d,0)$) {${}^\text{B}\bm{a}_{6,i1}^\textrm{T}$};
		\node(ba62)[bKreis] at ($(bg5) -(\d,1.5*\h)$) {${}^\text{B}\bm{a}_{6,iN}^\textrm{T}$};
		\node(bg4)[brechteck] at ($(ba61) -(\d, 0)$) {${}^\text{B}g_{4}^\textrm{T}$};
		\node(ba5)[bKreis] at ($(bg4) -(\d,0)$) {${}^\text{B}a_{5,i1}^\textrm{T}$};
		\node(bg3)[brechteck] at ($(ba5) -(\d, 0)$) {${}^\text{B}g_{3}^\textrm{T}$};
		\node(ba4)[bKreis] at ($(bg3) -(\d,0)$) {${}^\text{B}\bm{a}_{4,i1}^\textrm{T}$};
		\node(bg2)[brechteck] at ($(ba4) -(\d, 0)$) {${}^\text{B}g_{2}^\textrm{T}$};
		\node(ba2)[bKreis] at ($(bg2) -(\d,0)$) {${}^\text{B}\bm{a}_{2,i1}^\textrm{T}$};
		\node(bg0)[brechteck] at ($(ba2) -(\d, 0)$) {${}^\text{B}g_{0}^\textrm{T}$};
		\node(bat)[bKreis] at ($(bg0) -(\d,0)$) {${}^\text{B}\mathbf{\Theta}_{R}^\textrm{T}$};
		\draw[->, blue] (ba11)  -- (bg10);
		\draw[->, blue] (bg10)  -- (ba9);
		\draw[->, blue] (ba9)  -- (bg8);
		\draw[->, blue] (bg10)  -- (ba10);
		\draw[->, blue, dotted] (ba10)  -- (ba72);
		\draw[->, blue] (bg8)  -- (ba81);
		\draw[->, blue] (ba81)  -- (bg6);
		\draw[->, blue] (bg6)  -- (ba71);
		\draw[->, blue] (ba71) -- (bg5);
		\draw[->, blue] (ba72) -- (bg5);
		\draw[->, blue] (bg5) -- (ba61);
		\draw[->, blue, dotted] (bg5) -- (ba62);
		\draw[->, blue, dotted] (ba62) -- ($(bg0)- (0,1.5*\h)$);
		\draw[->, blue, dotted]  ($(bg0)- (0,1.5*\h)$) -- (bat);
		\draw[->, blue] (ba61) -- (bg4);
		\draw[->, blue] (bg4) -- (ba5);
		\draw[->, blue] (ba5) -- (bg3);
		\draw[->, blue] (bg3) -- (ba4);
		\draw[->, blue] (ba4) -- (bg2);
		\draw[->, blue] (bg2) -- (ba2);
		\draw[->, blue] (ba2) -- (bg0);
		\draw[->, blue] (bg0) -- (bat);
		
		\draw[->, blue] (tR) to[bend left] (bg0);
		\draw[->, blue] (a9) to[bend left] (bg10);
		\draw[->, blue] (a10) to[bend left] (bg10);
		\draw[->, blue] (a81) to[bend left] (bg8);
		\end{tikzpicture} }
	\caption{Partial forward (black) and backward (blue) pass for the parameter $\mathbf{\Theta}_{R}$ of a one-layered GATv2. Intermediate results in the forward pass are denoted by $\bm{a}_{l,ij}$ and ${}^\textrm{B}\bm{a}_{l,ij}$ in the backward pass, where $l$ indicates the layer. Functions are characterized by a rectangular shape and denoted as $g_p$, where $p$ is the number of the operation.}
	\label{fig:fbpass}
\end{figure}
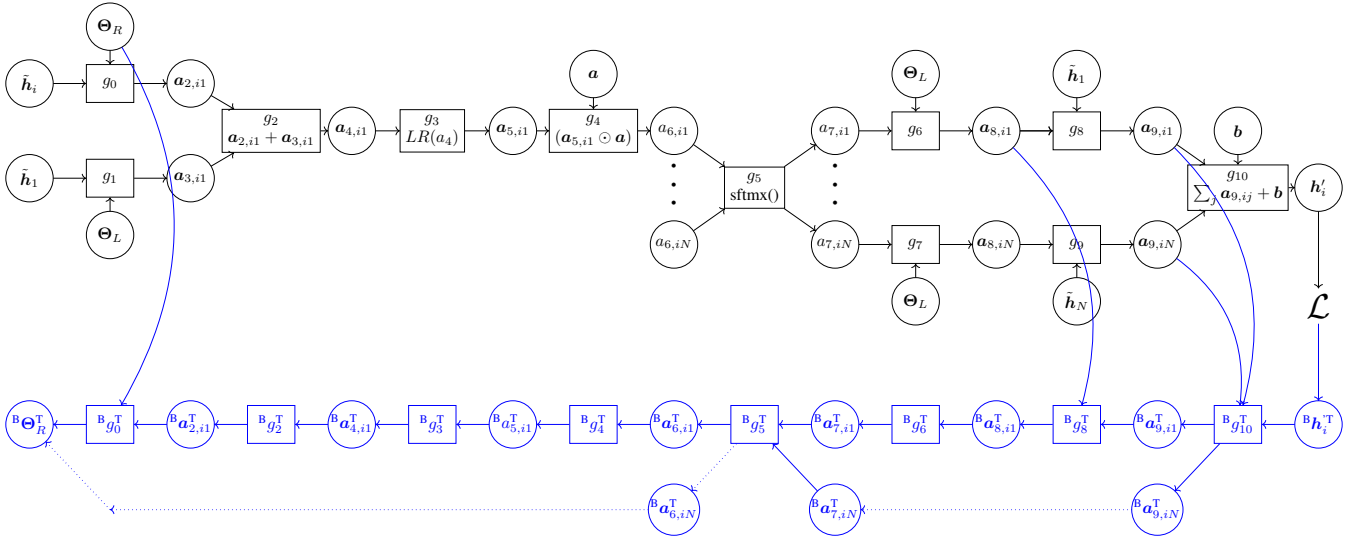

%% file: GradientBackpropGATv2.bbl
\begin{thebibliography}{1}
\providecommand{\url}[1]{#1}
\csname url@samestyle\endcsname
\providecommand{\newblock}{\relax}
\providecommand{\bibinfo}[2]{#2}
\providecommand{\BIBentrySTDinterwordspacing}{\spaceskip=0pt\relax}
\providecommand{\BIBentryALTinterwordstretchfactor}{4}
\providecommand{\BIBentryALTinterwordspacing}{\spaceskip=\fontdimen2\font plus
\BIBentryALTinterwordstretchfactor\fontdimen3\font minus
  \fontdimen4\font\relax}
\providecommand{\BIBforeignlanguage}[2]{{%
\expandafter\ifx\csname l@#1\endcsname\relax
\typeout{** WARNING: IEEEtranS.bst: No hyphenation pattern has been}%
\typeout{** loaded for the language `#1'. Using the pattern for}%
\typeout{** the default language instead.}%
\else
\language=\csname l@#1\endcsname
\fi
#2}}
\providecommand{\BIBdecl}{\relax}
\BIBdecl

\bibitem{understandGAT}
{Knyazev, Boris and Taylor, Graham W. and Amer, Mohamed R.}, ``Understanding
  attention and generalization in graph neural networks,'' in \emph{Proceedings
  of the 33rd International Conference on Neural Information Processing Systems
  (NIPS)}, 2019, vol.~33, pp. 4202--4212.

\bibitem{Neumeier.2023}
M.~Neumeier, A.~Tollk{\"u}hn, S.~Dorn, M.~Botsch, and W.~Utschick,
  ``Optimization and interpretability of graph attention networks for small
  sparse graph structures in automotive applications,'' in \emph{IEEE
  Intelligent Vehicles Symposium (IV)}, 2023.

\bibitem{pytorchGATv2CONV}
\BIBentryALTinterwordspacing
{PyTorch Geometric}, ``Gatv2conv.'' [Online]. Available:
  \url{https://pytorch-geometric.readthedocs.io/en/latest/modules/nn.html#torch_geometric.nn.conv.GATv2Conv}
\BIBentrySTDinterwordspacing

\bibitem{GATv2}
{Shaked Brody, Uri Alon, Eran Yahav}, ``How attentive are graph attention
  networks?'' in \emph{International Conference on Learning Representations
  (ICLR)}, 2022, vol.~10.

\bibitem{Velickovic.30.10.2017}
P.~Veli{\v{c}}kovi{\'c}, G.~Cucurull, A.~Casanova, A.~Romero, P.~Li{\`o}, and
  Y.~Bengio, ``Graph attention networks,'' in \emph{International Conference on
  Learning Representations (ICLR)}, 2018, vol.~6.

\end{thebibliography}
